\documentclass{article}
\usepackage{spconf,amsmath,graphicx}
\usepackage{times}
\usepackage{url}
\usepackage{latexsym}
\usepackage{tikz}
\usepackage{bm}
\usepackage{booktabs}
\usepackage{amssymb,amsmath}
\usepackage{relsize}
\usepackage[font={small}]{caption}

\newcommand{\newo}{GESD}
\newcommand{\newt}{AESD}

\newcommand{\RN}[1]{%
  \textup{\uppercase\expandafter{\romannumeral#1}}%
}
\usepackage[font={small}]{caption}
\newcounter{magicrownumbers}
\newcommand\rownumber{\stepcounter{magicrownumbers}\arabic{magicrownumbers}}

\title{Applying Deep Learning to Answer Selection: \\A Study and An Open Task}
\name{Minwei Feng, Bing Xiang, Michael R. Glass, Lidan Wang, Bowen Zhou}
\address{
   IBM Watson\\
  Yorktown Heights, NY, USA, 10598\\
  {\tt <mfeng|bingxia|mrglass|wangli|zhou>@us.ibm.com}
}
\begin{document}
%
\maketitle

\begin{abstract}
We apply a general deep learning framework to address the non-factoid question answering task.
Our approach does not rely on any linguistic tools and can be applied to different languages or domains. Various architectures are presented and compared. We create and release a QA corpus and setup a new QA task in the insurance domain. Experimental results demonstrate superior performance compared to the baseline methods and various technologies give further improvements. For this highly challenging task, the top-1 accuracy can reach up to 65.3\% on a test set, which indicates a great potential for practical use.
\end{abstract}

\section{Introduction}
\label{sec:intro}
Natural language understanding based spoken dialog system has been a popular topic in the past years of artificial intelligence renaissance. Many of those influential systems include a question answering module, e.g. Apple's Siri, IBM's Watson and Amazon's Echo.
In this paper, we address the Question Answering (QA) module in those spoken QA systems. We treat the QA from a text matching and selection perspective.
IBM's Watson system \cite{Ferrucci:journals-aim-Ferrucc10} is a classical example of the traditional way of doing Question Answering (QA). In this work we utilize a deep learning framework to accomplish the answer selection which is a key step in the QA task.
Hence QA is studied from an answer matching and selection perspective.
Given a question $q$ and an answer candidate pool $\{a_1, a_2, ... , a_s\}$ for that question ($s$ is the pool size), the goal is to find the best answer candidate $a_k$, $1 \leq k \leq s $~. If the selected answer $a_k$ is inside the ground truth set (one question could have more than one correct answer) of question $q$~, the question $q$ is considered to be answered correctly, otherwise it is answered incorrectly.
From the definition, the QA problem can be regarded as a binary classification problem. For each question, for each answer candidate, it may be appropriate or not. In order to find the best pair, we need a metric to measure the matching degree of each QA pair so that the QA pair with highest metric value will be chosen. 

The above definition is general. 
The only assumption made is that for every question there is an answer candidate pool. In practice, the pool can be easily built by using a general search engine like Google Search or an information retrieval software library like Apache Lucene. 

We created a data set by collecting question and answer pairs from the internet. All these question and answer pairs are in the insurance domain. The construction of this insurance domain QA corpus was driven by the intense scientific and commercial interest in this domain.
We released this corpus \footnote{git clone https://github.com/shuzi/insuranceQA.git} to create an open QA task, enabling other researchers to utilize it and supporting a fair comparison among different methods. The corpus consists of four parts: train, development, test1 and test2. Table~\ref{table:corpus_statistics} gives the data statistics. All experiments conducted in this paper are based on this corpus. To our best knowledge, it is the first time an insurance domain QA task has been released.

Our QA task requires specifying an answer candidate pool for each question in the development, test1 and test2 parts of the corpus. The released corpus contains totally 24\,981 unique answers. It is possible to use the whole answer space as the candidate pool, so that each question must be compared with 24\,981 answer candidates. However, this is impractical due to time consuming computations. In this paper, we set the pool size to be 500, so that it is both practical and still a challenging task. We put the ground truth answers into the pool and randomly sample negative answers from the answer space until the pool size reaches 500. 

The technology described in this paper with the released data set and benchmark task is targeting potential applications like online customer service. Hence it is not supposed to handle question answering tasks that require reasoning, e.g. is tomorrow Tuesday? (answer depends on if today is Monday.) The rest of the  paper is organized as follows: Sec. 2 describes the different architectures used this work; Sec. 3 provides the experimental setup details; experimental results and discussions are presented in Sec. 4; Sec. 5 contains related work and finally we draw conclusions in Sec. 6.

\begin{table}[t]
\small
\centering
\begin{tabular}{lrrr}
\toprule
                                                                                     &   \multicolumn{1}{l}{\textbf{Questions}} & \multicolumn{1}{r}{\textbf{Answers}}  & \multicolumn{1}{r}{\textbf{Question Word Count}} \\
\midrule
          \multicolumn{1}{l}{\textbf{Train}}                                        &   \multicolumn{1}{r}{$12\,887$}     & \multicolumn{1}{r}{$18\,540$}  & \multicolumn{1}{r}{$92\,095$}         \\
          \multicolumn{1}{l}{\textbf{Dev}}                                        &   \multicolumn{1}{r}{$1\,000$}     & \multicolumn{1}{r}{$1\,454$}  & \multicolumn{1}{r}{$7\,158$}           \\
          \multicolumn{1}{l}{\textbf{Test1}}                                        &   \multicolumn{1}{r}{$1\,800$}     & \multicolumn{1}{r}{$2\,616$}  & \multicolumn{1}{r}{$12\,893$}           \\
          \multicolumn{1}{l}{\textbf{Test2}}                                        &   \multicolumn{1}{r}{$1\,800$}     & \multicolumn{1}{r}{$2\,593$}  & \multicolumn{1}{r}{$12\,905$}           \\

\bottomrule
\end{tabular}
\caption{\small Corpus statistics: first two columns are the question and answer quantity; notice there could be multiple answers for some questions so that the answer quantity is larger than the question quantity; third column is the question total word count. The total number of answers is 24\,981 and the whole answer text contains 2\,386\,749 words}
\label{table:corpus_statistics}
\end{table}

\section{Model Description}
\label{sec:model}
In this section we describe the proposed deep learning framework and many variations based on that framework. However, the main idea of those different systems is the same: learn a distributed vector representation of a given question and its answer candidates and then use a similarity metric to measure the matching degree. We first developed two baseline systems for comparison.

\subsection{Baseline Systems} 
The first baseline system is a bag-of-words model. Step one is to train a word embedding by \cite{NIPS2013_Mikolov}. This word embedding provides the word vector for each token in the question and its candidate answer.  From these, the baseline system produces the idf-weighted sum of word vectors for the question and for all of its answer candidates. This produces a vector representation for the question and each answer candidate. The last step is to calculate the cosine similarity between each question/candidate pair. The pair with highest cosine similarity is returned as the answer.
The second baseline is an information retrieval (IR) baseline. The state-of-the-art weighted dependency model (WD) \cite{Bendersky:2010:LCI,Bendersky:2011:PCW} is used. The WD model employs a weighted combination of term-based and term proximity-based ranking features to score each candidate answer. Example features include counts of question bigrams in ordered and unordered windows of different sizes in each candidate answer, in addition to simple unigram counts. The basic idea is that important bigrams or unigrams in the question should receive higher weights when their frequencies are computed. Thus, the feature weights are assigned in accordance to the importance of the question unigrams or bigrams that they are defined over, where the importance factor is learned as part of the model training process.
Row 1 and 2 (first column \textbf{Idx}) of Table~\ref{table:experimental_results} are the baseline system results.

\subsection{CNNs-based System}
In this paper, a QA framework based on Convolutional Neural Networks (CNN) is presented. As summarized in Chapter 11 of \cite{Bengio-et-al-2015-Book}, a CNN leverages three important ideas that can help improve a machine learning system: \textbf{sparse interaction}, \textbf{parameter sharing} and \textbf{equivariant representation}.
\textit{Sparse interaction} contrasts with traditional neural networks where each output is interactive with each input. In a CNN, the filter size (or kernel size) is usually much smaller than the input size. As a result , the output is only interactive with a narrow window of the input. \textit{Parameter sharing} refers to reusing the filter parameters in the convolution operations, while the element in the weight matrix of traditional neural network will be used only once to calculate the output. 
\textit{Equivariant representation} is related to the idea of $k$-MaxPooling which is usually combined with a CNN. In this paper we always set $k=1$. So each filter of the CNN represents some feature, and after the convolution operation, the $1$-MaxPooling value represents the highest degree that the input contains the feature. The position of that feature in the input is irrelevant due to the convolution. This property is very useful for many NLP applications. Below is an example to demonstrate our CNN implementation.
\begin{equation}
\label{eq:input_filter}
\small
\left( \begin{array}{cccc}
w_{11} & w_{21} & w_{31} & w_{41} \\
w_{12} & w_{22} & w_{32} & w_{42} \\
w_{13} & w_{23} & w_{33} & w_{43} \end{array}\right)  \bigodot
 \left( \begin{array}{cc}
f_{11} & f_{21}  \\
f_{12} & f_{22}  \\
f_{13} & f_{23} \end{array}
 \right)
\end{equation}
The left matrix {\small$W$} is the input sentence. Each word is represented by a $3$-dimensional word embedding vector and the input length is $4$. The right matrix {\small$F$} represents the filter. The 2-gram filter size is $3 \times 2$~. The convolution output of the input {\small$W$} and the filter {\small$F$} is a $3$-dim vector {\small$O$}~, assuming zero padding has been done so that only a narrow convolution is conducted.
\begin{equation}
\label{eq:convout}
\small
\begin{aligned}
o_1 &= w_{11}f_{11} + w_{12}f_{12} + w_{13}f_{13} +w_{21}f_{21} + w_{22}f_{22} + w_{23}f_{23} \\ 
o_2 &= w_{21}f_{11} + w_{22}f_{12} + w_{23}f_{13} +w_{31}f_{21} + w_{32}f_{22} + w_{33}f_{23} \\
o_3 &= w_{31}f_{11} + w_{32}f_{12} + w_{33}f_{13} +w_{41}f_{21} + w_{42}f_{22} + w_{43}f_{23} \\
\end{aligned}
\end{equation}
After $1$-MaxPooling, the maximum of the 3 values will be kept for the filter {\small$F$} which indicates the highest degree that filter {\small$F$} matches the input {\small$W$}~.

\subsection{Training and Loss Function}
Different architectures will be described later. However all those different architectures share the same training and testing mechanism. In this paper we minimize a ranking loss similar to \cite{Weston-2014} \cite{NIPS2014_Hu}. During training, for each training question {\small$Q$} there is a positive answer {\small$A^+$}(the ground truth). A training instance is constructed by pairing this {\small$A^+$} with a negative answer {\small$A^-$}(a wrong answer) sampled from the whole answer space. The deep learning framework generates vector representations for the question and the two candidates: {\small$V_{Q}$}~, {\small$V_{A^+}$} and {\small$V_{A^-}$}~.  The cosine similarities {\small$cos(V_{Q},V_{A^+})$} and {\small$cos(V_{Q},V_{A^-})$} are calculated and the distance between the two similarities is compared to a margin: {\small$cos(V_{Q},V_{A^+}) - cos(V_{Q},V_{A^-}) < m$}~. $m$ is the margin. When this condition is satisfied, the implication is that the vector space embedding either ranks the positive answer below the negative answer, or does not sufficiently rank the positive answer above the negative answer. If {\small$cos(V_{Q},V_{A^+}) - cos(V_{Q},V_{A^-}) >= m$} there is no update to the parameters and a new negative example is sampled until the margin is less than $m$ (to reduce running time we set maximum 50 times in this paper). The hinge loss function is hence defined as follows:
\begin{equation}
\small
L = max\left\{0, m - cos(V_{Q},V_{A^+}) +cos(V_{Q},V_{A^-}) \right\}
\end{equation}
For testing, we calculate the {\small$cos(V_{Q},V_{candidate})$} between the question {\small$Q$} and each answer candidate {\small$V_{candidate}$} in the pool (size 500). The candidate answer with largest cosine similarity is selected.

\begin{figure}[t]
\begin{center}
\begin{tikzpicture}[scale=0.8]
  \begin{scope}
    \draw (0.0,0.0)rectangle (0.4,1.2) node[pos=.5] {\scriptsize A};  
    \draw (0.0,1.8)rectangle (0.4,3.0) node[pos=.5] {\scriptsize Q};   
    \draw (1.0,0.0)rectangle (2.0,1.2) node[pos=.5] {\scriptsize HL$_A$};   
    \draw (1.0,1.8)rectangle (2.0,3.0) node[pos=.5] {\scriptsize HL$_Q$};
    \draw (2.5,0.0)rectangle (3.5,1.2) node[pos=.5] {\scriptsize CNN$_A$};
    \draw (2.5,1.8)rectangle (3.5,3.0) node[pos=.5] {\scriptsize CNN$_Q$};
    \draw (4.0,0.0)rectangle (4.4,1.2) node[pos=.5] {\scriptsize \begin{tabular}{c} P \\ $+$ \\ T \end{tabular}};
    \draw (4.0,1.8)rectangle (4.4,3.0) node[pos=.5] {\scriptsize \begin{tabular}{c} P \\ $+$ \\ T \end{tabular}};
    
    \draw [->, thick] (0.4,0.6) -- (1.0,0.6);
    \draw [->, thick] (2.0,0.6) -- (2.5,0.6);
    \draw [->, thick] (0.4,2.4) -- (1.0,2.4);
    \draw [->, thick] (2.0,2.4) -- (2.5,2.4);
    \draw [->, thick] (3.5,0.6) -- (4.0,0.6);
    \draw [->, thick] (3.5,2.4) -- (4.0,2.4);
    \draw [->, thick] (4.4,0.6) -- (5.2,1.4);
    \draw [->, thick] (4.4,2.5) -- (5.2,1.8);
    \fill (5.5,1.6) circle[radius=1pt] ;
    \draw (5.5,1.6) circle[radius=10pt, thick];
    \node at (5.5,2.4) {\scriptsize Cosine};
    \node at (5.7,2.1) {\scriptsize Similarity};
  \end{scope}
\end{tikzpicture}
\end{center}
\caption{\small Architecture $\RN{1}$~. Q for question; A for answer; P is $1$-MaxPooling; T is $tanh$ layer; HL for hidden layer and HL already includes $tanh$ as its activation function.}
\label{fig:arc1}
\end{figure}
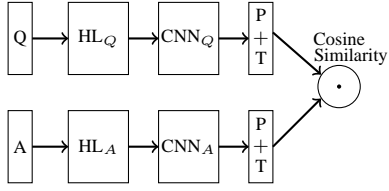

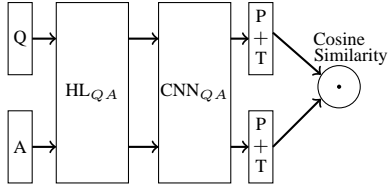
\begin{figure}[t]
\begin{center}
\begin{tikzpicture}[scale=0.8]
  \begin{scope}
    \draw (0.0,0.0)rectangle (0.4,1.2) node[pos=.5] {\scriptsize A};  
    \draw (0.0,1.8)rectangle (0.4,3.0) node[pos=.5] {\scriptsize Q};   
    \draw (0.8,0.0)rectangle (2.0,3.0) node[pos=.5] {\scriptsize HL$_{QA}$};   
    \draw (2.5,0.0)rectangle (3.7,3.0) node[pos=.5] {\scriptsize CNN$_{QA}$};
    \draw (4.0,0.0)rectangle (4.4,1.2) node[pos=.5] {\scriptsize \begin{tabular}{c} P \\ $+$ \\ T \end{tabular}};
    \draw (4.0,1.8)rectangle (4.4,3.0) node[pos=.5] {\scriptsize \begin{tabular}{c} P \\ $+$ \\ T \end{tabular}};
    
    \draw [->, thick] (0.4,0.6) -- (0.8,0.6);
    \draw [->, thick] (2.0,0.6) -- (2.5,0.6);
    \draw [->, thick] (0.4,2.4) -- (0.8,2.4);
    \draw [->, thick] (2.0,2.4) -- (2.5,2.4);
    \draw [->, thick] (3.7,0.6) -- (4.0,0.6);
    \draw [->, thick] (3.7,2.4) -- (4.0,2.4);
    \draw [->, thick] (4.4,0.6) -- (5.2,1.4);
    \draw [->, thick] (4.4,2.5) -- (5.2,1.8);
    \fill (5.5,1.6) circle[radius=1pt] ;
    \draw (5.5,1.6) circle[radius=10pt, thick];
    \node at (5.5,2.4) {\scriptsize Cosine};
    \node at (5.7,2.1) {\scriptsize Similarity};
  \end{scope}
\end{tikzpicture}
\end{center}
\caption{\small Architecture $\RN{2}$~.  QA means the weights of corresponding layer are shared by Q and A~.}
\label{fig:arc2}
\end{figure}

\begin{figure}[t]
\begin{center}

\begin{tikzpicture}[scale=0.8]
  \begin{scope}
    \draw (0.0,0.0)rectangle (0.4,1.2) node[pos=.5] {\scriptsize A};  
    \draw (0.0,1.8)rectangle (0.4,3.0) node[pos=.5] {\scriptsize Q};   
    \draw (0.8,0.0)rectangle (2.0,3.0) node[pos=.5] {\scriptsize HL$_{QA}$};   
    \draw (2.5,0.0)rectangle (3.7,3.0) node[pos=.5] {\scriptsize CNN$_{QA}$};
    \draw (4.8,0.0)rectangle (5.9,1.2) node[pos=.5] {\scriptsize HL$_A$};   
    \draw (4.8,1.8)rectangle (5.9,3.0) node[pos=.5] {\scriptsize HL$_Q$};
    \draw (4.0,0.0)rectangle (4.4,1.2) node[pos=.5] {\scriptsize \begin{tabular}{c} P \\ $+$ \\ T \end{tabular}};
    \draw (4.0,1.8)rectangle (4.4,3.0) node[pos=.5] {\scriptsize \begin{tabular}{c} P \\ $+$ \\ T \end{tabular}};

    \draw [->, thick] (0.4,0.6) -- (0.8,0.6);
    \draw [->, thick] (2.0,0.6) -- (2.5,0.6);
    \draw [->, thick] (0.4,2.4) -- (0.8,2.4);
    \draw [->, thick] (2.0,2.4) -- (2.5,2.4);
    \draw [->, thick] (3.7,0.6) -- (4.0,0.6);
    \draw [->, thick] (3.7,2.4) -- (4.0,2.4);
    \draw [->, thick] (4.4,0.6) -- (4.8,0.6);
    \draw [->, thick] (4.4,2.4) -- (4.8,2.4);
    \draw [->, thick] (5.9,0.6) -- (6.2,1.4);
    \draw [->, thick] (5.9,2.4) -- (6.2,1.8);
    \fill (6.5,1.6) circle[radius=1pt] ;
    \draw (6.5,1.6) circle[radius=10pt, thick];
    \node at (6.55,2.5) {\scriptsize Cosine};
    \node at (6.75,2.2) {\scriptsize Similarity};
  \end{scope}
\end{tikzpicture}
\end{center}
\caption{\small Architecture $\RN{3}$~. HL for hidden layer. Add another HL$_Q$ and HL$_A$ after CNN$_{QA}$~. }
\label{fig:arc3}
\end{figure}
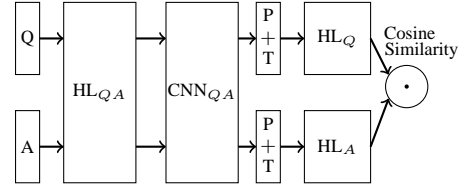

\begin{figure}[t]
\begin{center}

\begin{tikzpicture}[scale=0.8]

  \begin{scope}
    \draw (0.0,0.0)rectangle (0.4,1.2) node[pos=.5] {\scriptsize A};  
    \draw (0.0,1.8)rectangle (0.4,3.0) node[pos=.5] {\scriptsize Q};   
    \draw (0.8,0.0)rectangle (2.0,3.0) node[pos=.5] {\scriptsize HL$_{QA}$};   
    \draw (2.5,0.0)rectangle (3.7,3.0) node[pos=.5] {\scriptsize CNN$_{QA}$};
    \draw (4.8,0.0)rectangle (6.0,3.0) node[pos=.5] {\scriptsize HL$_{QA}$};   
    \draw (4.0,0.0)rectangle (4.4,1.2) node[pos=.5] {\scriptsize \begin{tabular}{c} P \\ $+$ \\ T \end{tabular}};
    \draw (4.0,1.8)rectangle (4.4,3.0) node[pos=.5] {\scriptsize \begin{tabular}{c} P \\ $+$ \\ T \end{tabular}};

    \draw [->, thick] (0.4,0.6) -- (0.8,0.6);
    \draw [->, thick] (2.0,0.6) -- (2.5,0.6);
    \draw [->, thick] (0.4,2.4) -- (0.8,2.4);
    \draw [->, thick] (2.0,2.4) -- (2.5,2.4);
    \draw [->, thick] (3.7,0.6) -- (4.0,0.6);
    \draw [->, thick] (3.7,2.4) -- (4.0,2.4);
    \draw [->, thick] (4.4,0.6) -- (4.8,0.6);
    \draw [->, thick] (4.4,2.4) -- (4.8,2.4);
    \draw [->, thick] (6.0,0.6) -- (6.3,1.4);
    \draw [->, thick] (6.0,2.4) -- (6.3,1.8);
    \fill (6.6,1.6) circle[radius=1pt] ;
    \draw (6.6,1.6) circle[radius=10pt, thick];
    \node at (6.6,2.5) {\scriptsize Cosine};
    \node at (6.8,2.2) {\scriptsize Similarity};
  \end{scope}
\end{tikzpicture}
\end{center}
\caption{\small Architecture $\RN{4}$~.  Add another shared hidden layer HL$_{QA}$ after CNN$_{QA}$~. }
\label{fig:arc4}
\end{figure}
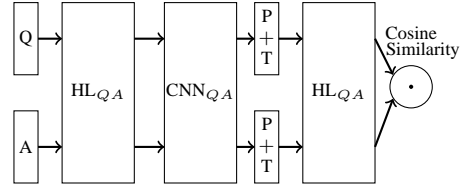

\begin{table*}[t]
\small
\centering
\begin{tabular}{lrrrl}
\toprule                                           
\multicolumn{1}{l}{\textbf{Idx}} &   \multicolumn{1}{r}{\textbf{Dev}} & \multicolumn{1}{r}{\textbf{Test1}}  & \multicolumn{1}{r}{\textbf{Test2}} & \multicolumn{1}{c}{\textbf{Description}}\\
\midrule               
\multicolumn{1}{l}{\rownumber}  &   \multicolumn{1}{r}{$31.9$}  & \multicolumn{1}{r}{$32.1$}  & \multicolumn{1}{r}{$32.2$} & Baseline: Bag-of-words \\
\multicolumn{1}{l}{\rownumber}  &   \multicolumn{1}{r}{$52.7$}  & \multicolumn{1}{r}{$55.1$}  & \multicolumn{1}{r}{$50.8$} & Baseline: metzler-bendersky IR model\\
\multicolumn{1}{l}{\rownumber}    &   \multicolumn{1}{r}{$44.2$}  & \multicolumn{1}{r}{$41.7$}  & \multicolumn{1}{r}{$39.5$} & Architecture $\RN{1}$:\hspace{0.2cm}  HL$_Q$(200) HL$_A$(200) CNN$_Q$(1000) CNN$_A$(1000) $1$-MaxPooling Tanh\\
\multicolumn{1}{l}{\rownumber}                                                                               &   \multicolumn{1}{r}{$58.2$}     & \multicolumn{1}{r}{$57.8$}  & \multicolumn{1}{r}{$53.6$} &Architecture $\RN{2}$:\hspace{0.1cm} HL$_{QA}$(200) CNN$_{QA}$(1000)$1$-MaxPooling Tanh  \\
 \multicolumn{1}{l}{\rownumber}                                                                               &   \multicolumn{1}{r}{$36.1$}  & \multicolumn{1}{r}{$33.6$}  & \multicolumn{1}{r}{$32.7$} &Architecture $\RN{3}$: HL$_{QA}$(200) CNN$_{QA}$(1000) HL$_Q$(1000) HL$_A$(1000) $1$-MaxPooling Tanh  \\
\multicolumn{1}{l}{\rownumber}                                                                              &   \multicolumn{1}{r}{$51.4$}     & \multicolumn{1}{r}{$50.5$}  & \multicolumn{1}{r}{$46.1$} &Architecture $\RN{4}$:   HL$_{QA}$(200) CNN$_{QA}$(1000)  HL$_{QA}$(1000) $1$-MaxPooling Tanh\\
\multicolumn{1}{l}{\rownumber}                                                  &   \multicolumn{1}{r}{$47.0$}     & \multicolumn{1}{r}{$46.7$}  & \multicolumn{1}{r}{$43.0$}  & Architecture $\RN{4}$:  HL$_{QA}$(200) CNN$_{QA}$(1000) HL$_{QA}$(500) $1$-MaxPooling Tanh \\
\multicolumn{1}{l}{\rownumber}                                                               &   \multicolumn{1}{r}{$60.6$}     & \multicolumn{1}{r}{$59.2$}  & \multicolumn{1}{r}{$55.1$}  & Architecture $\RN{2}$:\hspace{0.1cm}  HL$_{QA}$(200) CNN$_{QA}$(2000) $1$-MaxPooling Tanh   \\
\multicolumn{1}{l}{\rownumber}                                                           &   \multicolumn{1}{r}{$61.5$}     & \multicolumn{1}{r}{$61.3$}  & \multicolumn{1}{r}{$57.8$}  &Architecture $\RN{2}$:\hspace{0.1cm}  HL$_{QA}$(200) CNN$_{QA}$(3000) $1$-MaxPooling Tanh   \\
\multicolumn{1}{l}{\rownumber}                                                                 &   \multicolumn{1}{r}{$\bm{61.8}$}     & \multicolumn{1}{r}{ $\bm{62.8}$ }  & \multicolumn{1}{r}{$\bm{59.2}$}  &Architecture $\RN{2}$:\hspace{0.1cm}  HL$_{QA}$(200) CNN$_{QA}$(4000) $1$-MaxPooling Tanh (best result in this table)   \\
\multicolumn{1}{l}{\rownumber}                                                                             &   \multicolumn{1}{r}{$59.7$}     & \multicolumn{1}{r}{$59.3$}  & \multicolumn{1}{r}{$55.6$}& Architecture $\RN{5}$:\hspace{0.1cm} HL$_{QA}$(200) CNN$_{QA}$(1000) CNN$_{QA}$(1000) $1$-MaxPooling Tanh \\
\multicolumn{1}{l}{\rownumber}                                                                   &   \multicolumn{1}{r}{$59.9$}     & \multicolumn{1}{r}{$60.6$}  & \multicolumn{1}{r}{$55.9$}&Architecture $\RN{6}$: HL$_{QA}$(200) CNN$_{QA}$(1000) CNN$_{QA}$(1000) $1$-MaxPooling Tanh (2COST) \\
\multicolumn{1}{l}{\rownumber}                                                                               &   \multicolumn{1}{r}{$59.9$}     & \multicolumn{1}{r}{$58.7$}  & \multicolumn{1}{r}{$53.8$}&Architecture $\RN{2}$:\hspace{0.1cm} HL$_{QA}$(200) Augmented-CNN$_{QA}$(1000) $1$-MaxPooling Tanh  \\
\multicolumn{1}{l}{\rownumber}                                        &   \multicolumn{1}{r}{$60.0$}     & \multicolumn{1}{r}{$60.3$}  & \multicolumn{1}{r}{$54.3$}&Architecture $\RN{2}$:\hspace{0.1cm} HL$_{QA}$(200) Augmented-CNN$_{QA}$(2000) $1$-MaxPooling Tanh  \\
\multicolumn{1}{l}{\rownumber}                                                                    &   \multicolumn{1}{r}{$61.7$}     & \multicolumn{1}{r}{$62.2$}  & \multicolumn{1}{r}{$56.3$}&Architecture $\RN{2}$:\hspace{0.1cm} HL$_{QA}$(200) Augmented-CNN$_{QA}$(3000) $1$-MaxPooling Tanh  \\             
\bottomrule
\end{tabular}
\caption{\small Experimental Results. HL(200) means the hidden layer size is 200; CNN(1000) means there are 1000 filters used;  top one precision of Dev, Test1 and Test2 have been reported.}
\label{table:experimental_results}
\end{table*}

\subsection{Architectures}
In this subsection we demonstrate several proposed architectures for this QA task.
Figure~\ref{fig:arc1} shows the Architecture $\RN{1}$~. Q is the input question provided as input to the first hidden layer HL$_Q$.
The hidden layer (HL) is defined as $z = tanh(Wx + B)$. $W$ is the weight matrix; $B$ is the bias vector; $x$ is input; $z$ is the output of the activation function $tanh$. The output then flows to the CNN layer CNN$_{Q}$, applied to extract question side features. P is the MaxPooling layer (we always use $1$-MaxPooling in this paper) and T is the $tanh$ layer. Similar to the  question side, the answer A is processed by HL$_A$ and then features are extracted by CNN$_A$~. $1$-MaxPooling P and $tanh$ layer T will function in the last step. The result is a vector representation for both question and answer. The final output is the cosine similarity between these vectors. Row 3 of Table~\ref{table:experimental_results} is the  Architecture $\RN{1}$ result. 

Figure~\ref{fig:arc2} is the Architecture $\RN{2}$~. The main difference compared to Architecture $\RN{1}$ is that both question and answer sides share the same HL and CNN weights. Row 4 of Table~\ref{table:experimental_results} is the  Architecture $\RN{2}$ result.

We also consider architectures with a hidden layer after the CNN. Figure~\ref{fig:arc3} is the Architecture $\RN{3}$ in which another HL$_{Q}$ is added at the question side after the CNN and another HL$_A$ is added at the answer side after the CNN. Row 5 of Table~\ref{table:experimental_results} is the  Architecture $\RN{3}$ result. Architecture $\RN{4}$, shown in Figure~\ref{fig:arc4}, is similar except the second HL of both question and answer share the same HL$_{QA}$ weights. The rows 6 and 7 of Table~\ref{table:experimental_results} are the  Architecture $\RN{4}$ results.

\begin{figure}[t]
\begin{center}
\begin{tikzpicture}[scale=0.8]
  \begin{scope}
    \draw (0.0,0.0)rectangle (0.4,1.2) node[pos=.5] {\scriptsize A};  
    \draw (0.0,1.8)rectangle (0.4,3.0) node[pos=.5] {\scriptsize Q};   
    \draw (0.8,0.0)rectangle (2.0,3.0) node[pos=.5] {\scriptsize HL$_{QA}$};   
    \draw (2.5,0.0)rectangle (3.7,3.0) node[pos=.5] {\scriptsize CNN$_{QA}$};
    \draw (4.0,0.0)rectangle (5.2,3.0) node[pos=.5] {\scriptsize CNN$_{QA}$};   
    \draw (5.6,0.0)rectangle (6.0,1.2) node[pos=.5] {\scriptsize \begin{tabular}{c} P \\ $+$ \\ T \end{tabular}};
    \draw (5.6,1.8)rectangle (6.0,3.0) node[pos=.5] {\scriptsize \begin{tabular}{c} P \\ $+$ \\ T \end{tabular}};

    \draw [->, thick] (0.4,0.6) -- (0.8,0.6);
    \draw [->, thick] (2.0,0.6) -- (2.5,0.6);
    \draw [->, thick] (0.4,2.4) -- (0.8,2.4);
    \draw [->, thick] (2.0,2.4) -- (2.5,2.4);
    \draw [->, thick] (3.7,0.6) -- (4.0,0.6);
    \draw [->, thick] (3.7,2.4) -- (4.0,2.4);
    \draw [->, thick] (5.2,0.6) -- (5.6,0.6);
    \draw [->, thick] (5.2,2.4) -- (5.6,2.4);
    \draw [->, thick] (6.0,0.6) -- (6.3,1.4);
    \draw [->, thick] (6.0,2.4) -- (6.3,1.8);
    \fill (6.6,1.6) circle[radius=1pt] ;
    \draw (6.6,1.6) circle[radius=10pt, thick];
    \node at (6.6,2.5) {\scriptsize Cosine};
    \node at (6.8,2.2) {\scriptsize Similarity};
  \end{scope}
\end{tikzpicture}
\end{center}
\caption{\small Architecture $\RN{5}$~. Two shared CNN$_{QA}$~. }
\label{fig:arc5}
\end{figure}
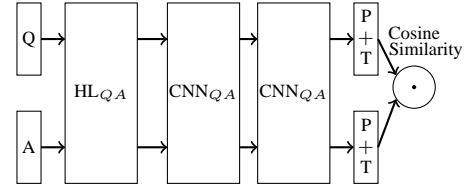

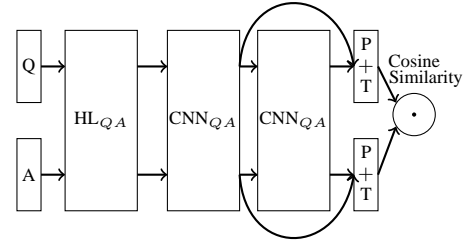
\begin{figure}[t]
\begin{center}
\begin{tikzpicture}[scale=0.8]
  \begin{scope}
    \draw (0.0,0.0)rectangle (0.4,1.2) node[pos=.5] {\scriptsize A};  
    \draw (0.0,1.8)rectangle (0.4,3.0) node[pos=.5] {\scriptsize Q};   
    \draw (0.8,0.0)rectangle (2.0,3.0) node[pos=.5] {\scriptsize HL$_{QA}$};   
    \draw (2.5,0.0)rectangle (3.7,3.0) node[pos=.5] {\scriptsize CNN$_{QA}$};
    \draw (4.0,0.0)rectangle (5.2,3.0) node[pos=.5] {\scriptsize CNN$_{QA}$};   
    \draw (5.6,0.0)rectangle (6.0,1.2) node[pos=.5] {\scriptsize \begin{tabular}{c} P \\ $+$ \\ T \end{tabular}};
    \draw (5.6,1.8)rectangle (6.0,3.0) node[pos=.5] {\scriptsize \begin{tabular}{c} P \\ $+$ \\ T \end{tabular}};
    
    \draw [->, thick] (3.7,0.6) .. controls (3.8,-0.8) and (5.35,-0.8) ..(5.6, 0.6);
    \draw [->, thick] (3.7,2.4) .. controls (3.8,3.8) and (5.35,3.8) ..(5.6, 2.4);
    \draw [->, thick] (0.4,0.6) -- (0.8,0.6);
    \draw [->, thick] (2.0,0.6) -- (2.5,0.6);
    \draw [->, thick] (0.4,2.4) -- (0.8,2.4);
    \draw [->, thick] (2.0,2.4) -- (2.5,2.4);
    \draw [->, thick] (3.7,0.6) -- (4.0,0.6);
    \draw [->, thick] (3.7,2.4) -- (4.0,2.4);
    \draw [->, thick] (5.2,0.6) -- (5.6,0.6);
    \draw [->, thick] (5.2,2.4) -- (5.6,2.4);
    \draw [->, thick] (6.0,0.6) -- (6.3,1.4);
    \draw [->, thick] (6.0,2.4) -- (6.3,1.8);
    \fill (6.6,1.6) circle[radius=1pt] ;
    \draw (6.6,1.6) circle[radius=10pt, thick];
    \node at (6.6,2.5) {\scriptsize Cosine};
    \node at (6.8,2.2) {\scriptsize Similarity};
  \end{scope}
\end{tikzpicture}
\end{center}
\caption{\small Architecture $\RN{6}$~. Two shared CNN$_{QA}$~. Two cost functions.}
\label{fig:arc6}
\end{figure}

Figure~\ref{fig:arc5} is the Architecture $\RN{5}$ where two layers of CNN$_{QA}$ are deployed. In section 2.2 we show the convolution output is a vector ($3$-dim in that example). This is true only for CNNs with a single filter. By applying multiple filters the result is a matrix. If there are 4 filters utilized for the example in section 2.2, the output is the following matrix.  
\begin{equation}
\small
\left( \begin{array}{ccc}
o_{11} & o_{21} & o_{31} \\
o_{12} & o_{22} & o_{32} \\
o_{13} & o_{23} & o_{33} \\
o_{14} & o_{24} & o_{34} \end{array}\right)
\end{equation}
Each row represents the output of one filter and each column represents a bigram of the input. This matrix is the input to the next CNN$_{QA}$ layer. For this second layer, every bigram is effectively one ``word'' and the previous filter's output for that bigram is its word embedding. 
Row 11 of Table~\ref{table:experimental_results} is the result for Architecture $\RN{5}$~.  Architecture $\RN{6}$ in Figure~\ref{fig:arc6} is similar to Architecture $\RN{5}$ except we utilize layer-wise supervision. After each CNN$_{QA}$ layer there is $1$-MaxPooling and a $tanh$ layer so that the cost function can be calculated and back-propagation can be conducted. The result of Architecture $\RN{6}$ is in row 12 of  Table~\ref{table:experimental_results}.

We have tried another three techniques to improve Architecture $\RN{2}$ in Figure~\ref{fig:arc2}~. First, the CNN filter quantity has been increased, see row 8 9 and 10 of Table~\ref{table:experimental_results}. Second, the convolution operation has been augmented to include skip-bigrams.
Consider the example in section 2.2, for the input and one filter in Eq.~\ref{eq:input_filter}, the augmented convolution operation will not only produce Eq.~\ref{eq:convout} but also the following discontinuous convolution:
\begin{equation}
\label{eq:convout_dis}
\scriptsize
\begin{aligned}
o_4 &= w_{11}f_{11} + w_{12}f_{12} + w_{13}f_{13} +w_{31}f_{21} + w_{32}f_{22} + w_{33}f_{23} \\ 
o_5 &= w_{21}f_{11} + w_{22}f_{12} + w_{23}f_{13} +w_{41}f_{21} + w_{42}f_{22} + w_{43}f_{23} \\
\end{aligned}
\end{equation}
The $1$-MaxPooling will still be applied to get the largest value among $[o_1, o_2, o_3, o_4, o_5]$ so that this filter is automatically adapted to match a bigram or skip-bigram feature. Rows 13 14 and 15 of Table~\ref{table:experimental_results} show the results. Third, we investigate the similarity metric. Until now, we have been using the cosine similarity which is widely adopted for vector space models. However, is cosine similarity the best option for this task? Table~\ref{table:experimental_results_metric} is the results for similarity metric study. Some metrics include hyperparameters and experiments with various hyperparameters have been conducted. We propose two novel metrics (\newo\ and \newt) which demonstrate superior performance.

\begin{table*}[t]
\small
\centering
\begin{tabular}{rrrl}
\toprule
\multicolumn{1}{r}{\textbf{Dev}} & \multicolumn{1}{r}{\textbf{Test1}}  & \multicolumn{1}{r}{\textbf{Test2}} & \multicolumn{1}{c}{\textbf{Description}}\\
\midrule    
\multicolumn{1}{r}{$58.2$}  & \multicolumn{1}{r}{$57.8$}  & \multicolumn{1}{r}{$53.6$} & cosine: $k(x, y)=\frac{xy^{\intercal}}{\|x\|\|y\|}$\\
\multicolumn{1}{r}{$58.5$}  & \multicolumn{1}{r}{$57.1$}  & \multicolumn{1}{r}{$53.3$} & polynomial: $k(x, y)= (\gamma xy^{\intercal} + c)^{d}$, $\gamma=0.5, d=2,c=1$ \\
\multicolumn{1}{r}{$56.8$}  & \multicolumn{1}{r}{$54.6$}  & \multicolumn{1}{r}{$52.6$} & polynomial: $k(x, y)= (\gamma xy^{\intercal} + c)^{d}$, $\gamma=1.0, d=2,c=1$ \\
\multicolumn{1}{r}{$55.0$}  & \multicolumn{1}{r}{$53.6$}  & \multicolumn{1}{r}{$48.2$} & polynomial: $k(x, y)= (\gamma xy^{\intercal} + c)^{d}$, $\gamma=1.5, d=2,c=1$ \\
\multicolumn{1}{r}{$57.1$}  & \multicolumn{1}{r}{$53.7$}  & \multicolumn{1}{r}{$51.5$} & polynomial: $k(x, y)= (\gamma xy^{\intercal} + c)^{d}$, $\gamma=0.5, d=3,c=1$ \\
\multicolumn{1}{r}{$55.3$}  & \multicolumn{1}{r}{$52.4$}  & \multicolumn{1}{r}{$48.7$} & polynomial: $k(x, y)= (\gamma xy^{\intercal} + c)^{d}$, $\gamma=1.0, d=3,c=1$ \\
\multicolumn{1}{r}{$52.5$}  & \multicolumn{1}{r}{$51.0$}  & \multicolumn{1}{r}{$47.2$} & polynomial: $k(x, y)= (\gamma xy^{\intercal} + c)^{d}$, $\gamma=1.5, d=3,c=1$ \\
\multicolumn{1}{r}{$61.3$}  & \multicolumn{1}{r}{$59.9$}  & \multicolumn{1}{r}{$57.0$} & sigmoid: $k(x, y)=tanh(\gamma xy^{\intercal} + c)$, $\gamma=0.5, c=1$ \\
\multicolumn{1}{r}{$61.6$}  & \multicolumn{1}{r}{$60.2$}  & \multicolumn{1}{r}{$57.1$} & sigmoid: $k(x, y)=tanh(\gamma xy^{\intercal} + c)$, $\gamma=1.0, c=1$ \\
\multicolumn{1}{r}{$60.2$}  & \multicolumn{1}{r}{$60.2$}  & \multicolumn{1}{r}{$55.7$} & sigmoid: $k(x, y)=tanh(\gamma xy^{\intercal} + c)$, $\gamma=1.5, c=1$ \\
\multicolumn{1}{r}{$60.0$}  & \multicolumn{1}{r}{$60.3$}  & \multicolumn{1}{r}{$54.7$} & RBF: $k(x, y)=exp(-\gamma \|x-y\|^{2})$, $\gamma=0.5$ \\
\multicolumn{1}{r}{$60.2$}  & \multicolumn{1}{r}{$57.0$}  & \multicolumn{1}{r}{$54.4$} & RBF: $k(x, y)=exp(-\gamma \|x-y\|^{2})$, $\gamma=1.0$ \\
\multicolumn{1}{r}{$58.4$}  & \multicolumn{1}{r}{$57.3$}  & \multicolumn{1}{r}{$53.8$} & RBF: $k(x, y)=exp(-\gamma \|x-y\|^{2})$, $\gamma=1.5$ \\
\multicolumn{1}{r}{$60.8$}  & \multicolumn{1}{r}{$60.3$}  & \multicolumn{1}{r}{$57.0$} & euclidean: $k(x, y)=\frac{1}{1+\|x-y\|}$ \\
\multicolumn{1}{r}{$42.2$}  & \multicolumn{1}{r}{$42.5$}  & \multicolumn{1}{r}{$38.2$} & exponential: $k(x, y)=exp(-\gamma \|x-y\|_{1})$, $\gamma=0.5$ \\
\multicolumn{1}{r}{$41.4$}  & \multicolumn{1}{r}{$39.5$}  & \multicolumn{1}{r}{$36.0$} & exponential: $k(x, y)=exp(-\gamma \|x-y\|_{1})$, $\gamma=1.0$ \\
\multicolumn{1}{r}{$48.2$}  & \multicolumn{1}{r}{$45.1$}  & \multicolumn{1}{r}{$41.6$} & exponential: $k(x, y)=exp(-\gamma \|x-y\|_{1})$, $\gamma=1.5$ \\
\multicolumn{1}{r}{$51.0$}  & \multicolumn{1}{r}{$49.5$}  & \multicolumn{1}{r}{$46.4$} & manhattan: $k(x, y)=\frac{1}{1+\|x-y\|_{1}}$ \\
\multicolumn{1}{r}{$62.5$}  & \multicolumn{1}{r}{$61.4$}  & \multicolumn{1}{r}{$59.0$} & \newo: $k(x, y)=\frac{1}{1+\|x-y\|}\cdot\frac{1}{1+exp(-\gamma(xy^{\intercal} + c))}$, $\gamma=0.5,c=1$ \\
\multicolumn{1}{r}{$62.9$}  & \multicolumn{1}{r}{$62.1$}  & \multicolumn{1}{r}{$59.3$} & \newo: $k(x, y)=\frac{1}{1+\|x-y\|}\cdot\frac{1}{1+exp(-\gamma(xy^{\intercal} + c))}$, $\gamma=1.0,c=1$ \\
\multicolumn{1}{r}{$62.6$}  & \multicolumn{1}{r}{$62.1$}  & \multicolumn{1}{r}{$59.2$} & \newo: $k(x, y)=\frac{1}{1+\|x-y\|}\cdot\frac{1}{1+exp(-\gamma(xy^{\intercal} + c))}$, $\gamma=1.5,c=1$ \\
\multicolumn{1}{r}{$63.1$}  & \multicolumn{1}{r}{$61.9$}  & \multicolumn{1}{r}{$58.2$} & \newt: $k(x, y)=\frac{0.5}{1+\|x-y\|}{+}\frac{0.5}{1+exp(-\gamma(xy^{\intercal} + c))}$, $\gamma=0.5,c=1$ \\
\multicolumn{1}{r}{$63.4$}  & \multicolumn{1}{r}{$61.7$}  & \multicolumn{1}{r}{$58.7$} & \newt: $k(x, y)=\frac{0.5}{1+\|x-y\|}+\frac{0.5}{1+exp(-\gamma(xy^{\intercal} + c))}$, $\gamma=1.0,c=1$ \\
\multicolumn{1}{r}{$62.8$}  & \multicolumn{1}{r}{$62.0$}  & \multicolumn{1}{r}{$57.7$} & \newt: $k(x, y)=\frac{0.5}{1+\|x-y\|}+\frac{0.5}{1+exp(-\gamma(xy^{\intercal} + c))}$, $\gamma=1.5,c=1$ \\
\midrule

\multicolumn{1}{r}{$63.5$}  & \multicolumn{1}{r}{$62.5$}  & \multicolumn{1}{r}{$60.2$} & \newo: $k(x, y)=\frac{1}{1+\|x-y\|}\cdot\frac{1}{1+exp(-\gamma(xy^{\intercal} + c))}$, $\gamma=1.0$, ~2000 filters \\
\multicolumn{1}{r}{$64.3$}  & \multicolumn{1}{r}{$65.1$}  & \multicolumn{1}{r}{$61.0$} & \newo: $k(x, y)=\frac{1}{1+\|x-y\|}\cdot\frac{1}{1+exp(-\gamma(xy^{\intercal} + c))}$,  $\gamma=1.0$, ~3000 filters \\
\multicolumn{1}{r}{$\bm{65.4}$}  & \multicolumn{1}{r}{$\bm{65.3}$}  & \multicolumn{1}{r}{$61.0$} & \newo: $k(x, y)=\frac{1}{1+\|x-y\|}\cdot\frac{1}{1+exp(-\gamma(xy^{\intercal} + c))}$,  $\gamma=1.0$, ~4000 filters \\
\multicolumn{1}{r}{$64.5$}  & \multicolumn{1}{r}{$62.7$}  & \multicolumn{1}{r}{$60.1$} & \newt: $k(x, y)=\frac{0.5}{1+\|x-y\|}+\frac{0.5}{1+exp(-\gamma(xy^{\intercal} + c))}$, $\gamma=1.0$, ~2000 filters \\
\multicolumn{1}{r}{$64.3$}  & \multicolumn{1}{r}{$63.3$}  & \multicolumn{1}{r}{$\bm{62.2}$} & \newt: $k(x, y)=\frac{0.5}{1+\|x-y\|}+\frac{0.5}{1+exp(-\gamma(xy^{\intercal} + c))}$, $\gamma=1.0$, ~3000 filters \\
\multicolumn{1}{r}{$63.9$}  & \multicolumn{1}{r}{$64.5$}  & \multicolumn{1}{r}{$61.1$} & \newt: $k(x, y)=\frac{0.5}{1+\|x-y\|}+\frac{0.5}{1+exp(-\gamma(xy^{\intercal} + c))}$, $\gamma=1.0$, ~4000 filters \\

 \bottomrule
\end{tabular}
\caption{\small Experimental results of various similarities. All results in above part are based on Architecture $\RN{2}$ with 1000 filters (corresponding to Row 4 in Table~\ref{table:experimental_results}).
In the bottom part, the results are based on Architecture $\RN{2}$ using the proposed metric with more filters. $k(x,y)$ is the similarity between vector $x$ and $y$. $\|x\|$ is the $L_2$ norm and $\|x\|_1$ is the $L_1$ norm. $xy^{\intercal}$ represents the inner product of $x$ and $y$. We always normalize the question and answer vectors before calculating the similarity. Highest number in each column is in \textbf{bold} font.}
\label{table:experimental_results_metric}
\end{table*}

\section{Experimental Setup}
The deep learning framework in this paper has been built from scratch using Java. To improve speed, we adopt the HOGWILD approach \cite{NIPS2011_hogwild}~. Each thread processes one training instance at one time and updates the weights of the neural networks. There is no locking in any thread. The word embedding (100 dimensions) is trained by word2vec \cite{NIPS2013_Mikolov} and used for initialization. Word embeddings are also parameters and are optimized for the QA task. Stochastic Gradient Descent is the optimization strategy and the L2-norm is also added in the loss function. In this paper, the weight of the L2-norm is $0.0001$, the learning rate is $0.01$ and margin $m$ is $0.009$~. Those hyperparameters are chosen based on previous experiences in using deep learning on this data and they are not very sensitive within reasonable range. The utilized computing resources for this work are enormous. We heavily occupy a Power 7 cluster which consists of 75 machines. Each machine has 32 physical cores and each core supports 2-4  hyperthreading. The HOGWILD approach will bring some randomness due to no locking. Even with locking, the thread scheduler would alter the order of examples between runs so that randomness would still exist.
Therefore, for each row in Table~\ref{table:experimental_results} (except for row 1 2) and Table~\ref{table:experimental_results_metric}, 10 experiments have been conducted on the dev set and the run with best dev score is chosen to calculate the test scores.

\section{Results and Discussions}
In this section, detailed analysis on experimental results are given. From Table~\ref{table:experimental_results} and \ref{table:experimental_results_metric} the following conclusions can be made: \textbf{(1)} baseline 1 only utilizes word embeddings and baseline 2 is based on traditional term based features. Our proposed method can reach significantly better accuracy which demonstrates the superiority of deep learning approach; \textbf{(2)} using separate hidden layer (HL) or CNN layers for Q and A has worse performance compared to a shared HL or CNN layer (Table~\ref{table:experimental_results}, row 3 vs. 4, row 5 vs. 6). This is reasonable because for a shared layer network, the corresponding elements in Q and A vector are guaranteed to represent the same CNN filter convolution result while for network with separate Q and A layers, there is no such constraint and the optimizer has to learn over a set of double sized parameters. Hence the optimizer faces greater difficulty;   \textbf{(3)} adding a HL after the CNN degrades the performance (Table~\ref{table:experimental_results}, row 4 vs. 6 and 7). This proves that CNN already captures useful features for QA matching and unnecessary mapping the features to another space makes no sense at all; \textbf{(4)} increasing the CNN filter quantity can capture more features which gives notable improvement (Table~\ref{table:experimental_results}, row 4 vs. 8, 9 and 10); \textbf{(5)} two layers of CNN can represent a higher level of abstraction with wider range in the input. Hence going deeper by using two CNN layers improves the accuracy (Table~\ref{table:experimental_results}, row 4 vs. 11); \textbf{(6)} effective learning in deep networks is often a difficult task. Layer-wise supervision can alleviate the problem (Table~\ref{table:experimental_results}, row 11 vs. 12); \textbf{(7)} combining bigram and skip-bigram features brings gain on Test1 but not on Test2 (Table~\ref{table:experimental_results}, row 4 vs. 13, row 8 vs. 14, row 9 vs. 15); \textbf{(8)} Table~\ref{table:experimental_results_metric} shows that with the same model capacity, similarity metric plays an important role and the widely used cosine similarity is not the best choice for this task. The similarity in Table~\ref{table:experimental_results_metric} can be categorized into three classes: $L1$-norm based metric which is the semantic distance of Q and A summed from each coordinate axis; $L2$-norm based metric which is the straight-line semantic distance of Q and A; inner product based metric which measures the angle between Q and A. We propose two new metrics that combine $L2$-norm and inner product by multiplication (\newo\ Geometric mean of Euclidean and Sigmoid Dot product) and addition (\newt\ Arithmetic mean of Euclidean and Sigmoid Dot product). The proposed two metrics are the best among all compared metrics. Finally, in the bottom of Table~\ref{table:experimental_results_metric} it is clear that with more filters, the proposed metric can achieve even better performance.

\section{Related Work}
Deep learning technology has been widely used in machine learning tasks, often demonstrating superior performance compared to traditional methods. Many of those applications focus on classification related tasks, e.g. on image recognition \cite{lecun-01a}, on speech \cite{Schwenk:2007:CSL} \cite{export:171498} \cite{DBLP:conf/icassp/GravesMH13} and on machine translation \cite{devlin-EtAl:2014} \cite{sundermeyer-EtAl:2014:EMNLP2014}. This paper is based on many prior works on utilizing deep learning for NLP tasks: Gao et al. \cite{gao-EtAl:2014:EMNLP2014} proposed a CNN based network which maps source-target document pairs to embedding vectors such that
the distance between source documents
and their corresponding interesting
targets is minimized. 
Lu and Li \cite{Lu_NIPS2013_5019} propose a CNN based deep network for a short text matching task; Hu et al. \cite{NIPS2014_Hu} also use several CNN based networks for sentence matching;
Kalchbrenner et al. \cite{Kalchbrenner14} use a CNN for sentiment prediction and question classification; Kim \cite{kim:2014:EMNLP2014} uses a CNN in sentiment analysis; Zeng et al. \cite{zeng-EtAl:2014:Coling} use a CNN for relation classification; Socher et al. \cite{SocherEtAl2011:RNN} \cite{SocherEtAl2011:PoolRAE} use a recursive network for paraphrase detection and parsing; Iyyer et al. \cite{Iyyer:Boyd-Graber:Claudino:Socher:Daume-2014} propose a recursive network for factoid question answering; Weston et al. \cite{Weston-2014} use a CNN for hashtag prediction;
Yu et al. \cite{Yu:2014} use a CNN for answer selection;
Yin and Sch\"{u}tze \cite{yin-schutze:2015:NAACL-HLT1} use a bi-CNN for paraphrase identification. Our work follows the spirit of many previous work in the sense that we utilize CNN to map natural language sentences into embedding vectors so that the similarity can be calculated. However this paper has conducted extensive experiments over various architectures which are not included in previous work. Furthermore, we explored different similarity metrics, skip-bigram based convolution and layerwise supervision which have not been presented in previous work.

\section{Conclusions}
In this paper, the spoken question answering system is studied from an answer selection perspective by employing a deep learning framework. The proposed framework does not rely on any linguistic tool and can be easily adapted to different languages or domains. Our work serves as solid evidence that deep learning based QA is an encouraging research direction. The scientific contributions can be summarized as follows: \textbf{(1)} creating a new QA task in the insurance domain and releasing a new corpus so that different methods can be fairly compared; \textbf{(2)} proposing a general deep learning framework with several variants for the QA task and comparison experiments have been conducted; \textbf{(3)} utilizing novel techniques that bring improvements: multi-layer CNN with layer-wise supervision, augmented CNN with discontinuous convolution and novel similarity metric that combine both $L2$-norm and inner product information; \textbf{(4)} the best scores in this paper are very promising: for this challenging task (select one answer from a pool with size 500), the top one accuracy of test corpus can reach up to 65.3\%; \textbf{(5)} for researchers who want to proceed with this task, this paper provides valuable guidance: a shared layer structure should be adopted; no need to append a hidden layer after the CNN; two levels of CNN with layer-wise training improves accuracy; discontinuous convolution sometimes can help; the similarity metric plays a crucial role and the proposed metric is preferred and finally increasing the filter quantity brings improvement.


\newpage
\bibliographystyle{IEEEbib}
\bibliography{Template}

\end{document}